\documentclass{fairmeta}
\usepackage[most]{tcolorbox}

\usepackage{amsmath}
\usepackage{amssymb}
\usepackage{booktabs}
\usepackage{multirow}
\usepackage{graphicx}
\usepackage{tcolorbox}
\usepackage{makecell}
\usepackage{subcaption}
\usepackage{subcaption}
\usepackage{xcolor} 
\usepackage{amsmath}
\usepackage{amssymb}
\usepackage[export]{adjustbox}
\usepackage{wrapfig}
\usepackage{hyperref}

\newcommand{\m}{GIVE}
\renewcommand{\cite}[1]{\citep{#1}}
\title{\m{}: Grounding Human Gestures in Vision-Language-Action Models}

\author[1, *]{Pengfei Liu}
\author[1, *]{Gen Li}
\author[1]{Junqiao Fan}
\author[1]{Boyu Ma}
\author[1]{Jindou Jia}
\author[1]{Yang Xiao}
\author[1, \dagger]{Jianfei Yang}

\affiliation[1]{MARS Lab, Nanyang Technological University}

\contribution[*]{Equal contribution}
\contribution[\dagger]{Corresponding author}
\abstract{
Human communication is inherently multimodal, where language is often accompanied by non-verbal cues such as gestures to convey intentions. However, current Vision-Language-Action (VLA) models treat robotic manipulation as a pure text-driven task, overlooking the important role of gestures in Human-Robot Interaction (HRI). 
This often leads to inaccurate intent grounding and unreliable manipulation when language instructions are ambiguous or underspecified.
To address this challenge, we propose \m{} (\textbf{G}esture \textbf{I}ntent via \textbf{V}isual-Semantic \textbf{E}nhancement), an effective approach that enhances pre-trained VLA models with human gesture understanding without architectural modifications. Specifically, GIVE incorporates gesture information through two complementary pathways: a visual pathway that overlays hand skeletons and fingertip rays onto robot observations for explicit object grounding, and a semantic pathway that generates high-level descriptions of human gestures and task instructions for robust intent grounding. 
By jointly leveraging visual and semantic guidance, \m{} enables VLA policies to better associate gestures with manipulation behaviors and adapt to dynamic interaction intents.
In real-world HRI experiments, \m{} substantially outperforms the baseline, improving target object recognition accuracy by 40\% and overall task success rate by 80\%, while demonstrating strong robustness and generalization to unseen spatial layouts and diverse participants.
}

\correspondence{\email{jianfei.yang@ntu.edu.sg}, \email{pengfei007@e.ntu.edu.sg}}

\metadata[Project site]{\url{https://luis-cloud-sg.github.io/GIVE-project/}}

\begin{document}

\maketitle
\section{Introduction}

Enabling robots to seamlessly interact with humans and comprehend their intentions is a fundamental pursuit of embodied intelligence~\cite{hou2026world,goodrich2008human,yifan2025embodied}. In natural Human-Robot Interaction (HRI), communication is inherently multimodal.
Rather than relying solely on verbal commands, humans routinely complement language with a rich set of non-verbal cues, including gaze, posture, and gestures, to express intentions in a more intuitive, efficient, and contactless manner~\cite{gaze2act, posturehri, gesturehri}. 
Among these cues, human gestures are especially important as they provide a direct and physically grounded channel for intention expression. For example, a pointing gesture can instantly specify a target object, while an open palm can signal a handover request, effectively clarifying interaction intents within the physical workspace~\cite{strabala2013towards,cakmak2011human}.

\begin{figure}[!t]
\centering
\includegraphics[width=\textwidth]{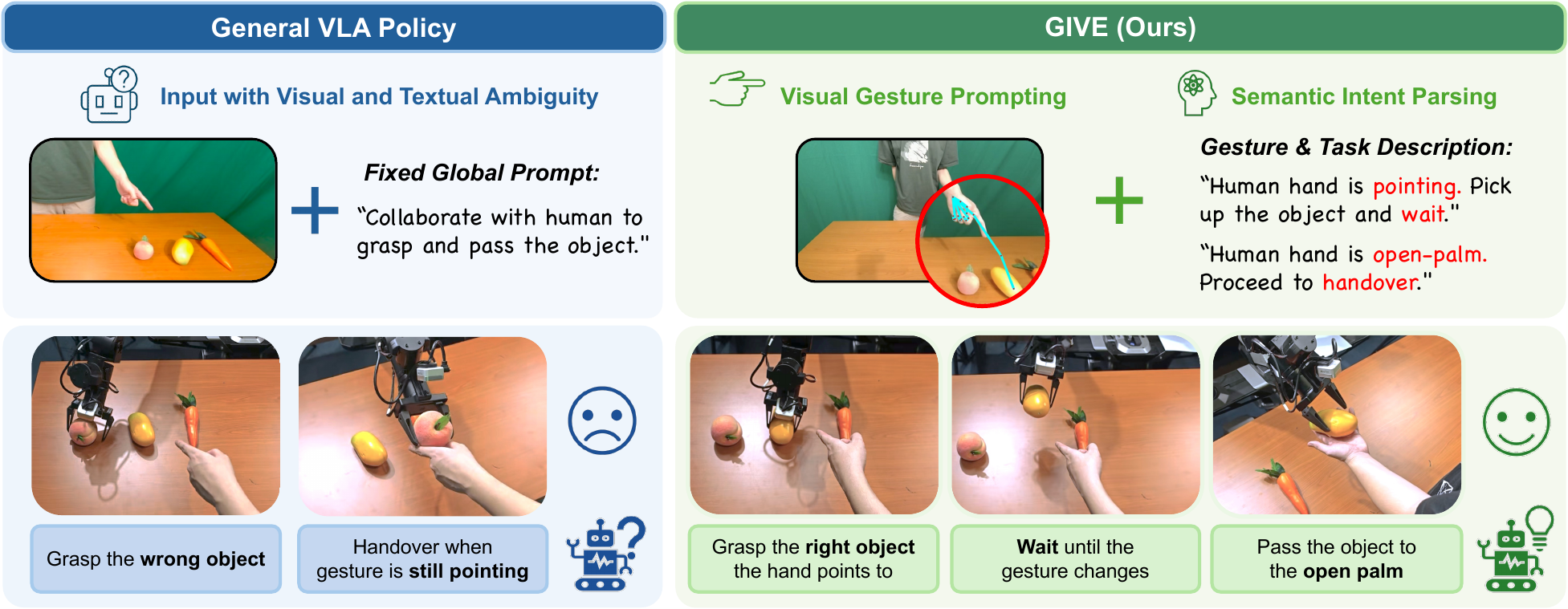}
\caption{\textbf{Comparison between VLA policies and \m{}}. General VLA policies struggle to accurately ground target objects and interpret human intent from hand gestures, whereas \m{} incorporates visual-semantic guidance that improves intent comprehension and enhances HRI performance.}
\label{fig:teaser}
\end{figure}

Despite the vital role of such intuitive non-verbal cues, Vision-Language-Action (VLA) models, which have recently emerged as a powerful paradigm, predominantly treat robotic manipulation as a purely language-driven task~\cite{brohan2022rt, intelligence2025pi_,zitkovich2023rt,o2024open,kim2024openvla,team2024octo}. 
However, conditioning exclusively on language directly contradicts the multimodal nature of human interaction.
Empirical studies in social science demonstrate that verbal language alone carries only about 30\% to 35\% of the meaning in face-to-face interactions, with the remaining 65\% to 70\% being conveyed through non-verbal channels such as gestures and body movements~\cite{birdwhistell2010kinesics}. 
Furthermore, cognitive linguistics reveals that speech and gestures are tightly coupled and jointly reflect an underlying intention representation during communication~\cite{mcneill1992hand}.
Therefore, relying solely on text to infer human intent is inherently insufficient, as described in Fig.~\ref{fig:teaser} (left), making it difficult for VLA models to associate physical gestures with their corresponding intentions.
This limitation becomes particularly critical when verbal commands are ambiguous or underspecified, often leading to intent misinterpretations and execution failures during robotic manipulation.

To address this limitation, we argue that effective HRI requires moving beyond text-conditioned policies by incorporating human gesture information as an additional source of interaction guidance. 
Instead of directly encoding geometric representations into policy networks, we propose to transform gestures into visual and semantic guidance that augments the robot’s observation space, enabling better grounding of human intent in manipulation behaviors.
Our key insight is that these two modalities play complementary roles: visual cues provide precise object grounding, while semantic cues capture high-level intent transitions driven by gesture changes.
Based on this insight, we propose \m{} (\textbf{G}esture \textbf{I}ntent via \textbf{V}isual-Semantic \textbf{E}nhancement), an approach that seamlessly integrates human gesture signals into pre-trained VLA models to enhance HRI.
Notably, our approach does not introduce additional learnable modules or architectural modifications to the original policy network.
Instead, gesture information is injected through visual prompting and semantic parsing, as illustrated in Fig.~\ref{fig:teaser} (right).
In the visual pathway, a 2D hand pose estimation module extracts hand joint keypoints and overlays the skeleton directly onto the robot's visual observation. 
For pointing gestures, we further construct a continuous fingertip-directed ray on the image plane, explicitly indicating the intended interaction object. 
In the semantic pathway, a pretrained Vision-Language Model (VLM) interprets the scene from a global view and generates textual descriptions of both human gestures and task instructions, providing high-level contextual understanding. 
By combining semantic and visual guidance, the policy can better ground human intent and adapt its behavior across gesture variations, enabling more reliable human-robot interaction.
Compared with the baseline model, \m{} improves the target object recognition accuracy by 40\% and boosts overall handover success rate by 80\%.

    In summary, our contributions are threefold: (1) we propose \m{}, a simple yet effective method that \textbf{injects human gesture cues into pre-trained VLA models without modifying their architecture}, enabling effective human intent grounding; (2) we introduce \textbf{a dual-path gesture guidance strategy} that transforms human gestures into complementary semantic and visual guidance, providing structured multimodal conditioning for robust action generation; and (3) we present \textbf{extensive real-world experiments} on a Galaxea R1-Lite dual-arm robot, showing consistent improvements over baselines, while demonstrating robustness and generalization across unseen spatial layouts and human participants.
\section{Related Work}
\label{sec:related_work}

\textbf{Vision-Language-Action Models.} The rapid development of foundation models has driven the emergence of VLAs, which directly map language instructions and visual observations to low-level motor commands~\cite{brohan2022rt, intelligence2025pi_}. Within this paradigm, RT-2 demonstrates the successful transfer of semantic reasoning from large VLMs to robotic control~\cite{zitkovich2023rt}, while the Open X-Embodiment initiative promotes cross-platform generalist policies through multi-embodiment data integration~\cite{o2024open}. Building on this, OpenVLA provides a scaled, open-source architecture trained on real-world demonstrations~\cite{kim2024openvla}, and $\pi_{0.5}$ further enhances open-world generalization via heterogeneous co-training across diverse data domains~\cite{intelligence2025pi_}. Despite these advancements, recent VLA approaches~\cite{lin2025evo0,lin2025evo1,liu2026palm,lin2026evo} remain predominantly designed for static, instruction-driven tasks. They struggle to dynamically deduce implicit human intents or capture subtle intent transitions from RGB observations, leading to limited coordination performance in unstructured HRI scenarios.

\textbf{Gesture-Based Intent Grounding.} Human gestures provide a direct and physically grounded channel for intent expression in robotic interaction, complementing language instructions in object referral and interaction-state representation. Existing approaches mainly follow two paradigms: vision-based prompting and text-based reasoning. Vision-based prompting approaches rely on hand keypoints, pointing rays, segmentation, or target masks for spatial grounding~\cite{affgrasp,ellmer,moon2020i2l, edge2023diver, hu2022augmented}, but often lack semantic understanding of task phases. RoboNurse-VLA applies VLA models to robotic handover by combining visual segmentation with explicit voice commands, but does not explicitly model gesture-driven intent transitions~\cite{li2025robonurse}. Text-based reasoning approaches use Large Language Models (LLMs) or VLMs for intention reasoning~\cite{kobzarev2025gestllm, lin2023gesture, chen2025intentionvla}, but lack precise spatial anchoring between gestures and physical objects. IntentionVLA studies embodied intention understanding through text-based semantic reasoning, but does not directly ground dynamic spatial gestures to robot manipulation behaviors~\cite{chen2025intentionvla}. In contrast, our method combines gesture-aware visual prompts with semantic intent descriptions, enabling pre-trained VLA policies to achieve both object-level spatial grounding and phase-level intent understanding.

\section{Method}
\label{sec:method}

\subsection{Overview}
\label{subsec:task_formulation}

\textbf{Problem Formulation.} We focus on a representative grasp-then-handover HRI task involving multiple interaction stages, including target indication, object grasping, and object delivery.
At each timestep $t$, the system receives multi-view visual observations $\mathcal{O}_t = \{o_t^{global}, o_t^{wrist}\}$ from the head and wrist cameras, along with a language instruction $l_t$. A control policy $\pi_\theta(a_t \mid \mathcal{O}_{t}, l_t)$ is learned to output an action sequence $a_t \in \mathcal{A}$, which consists of a continuous 6-DoF end-effector pose and a 1-D gripper command.

\begin{figure}[!t]
\centering
\includegraphics[width=\textwidth]{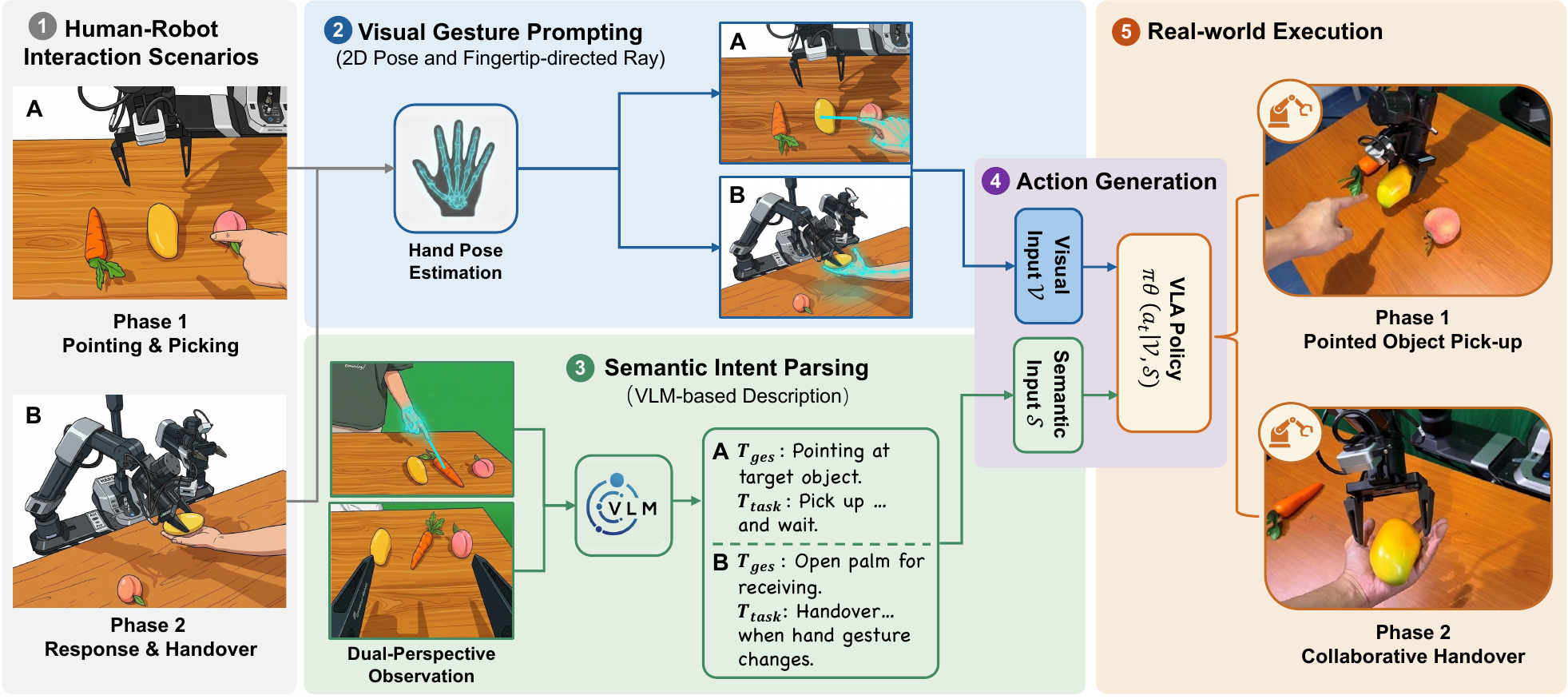}
\caption{\textbf{\m{} for VLA models in HRI scenarios.} \m{} transforms human gestures into two complementary guidance pathways: visual gesture prompting provides spatial grounding of manipulation targets, while semantic intent parsing captures high-level intent description. These cues are fed into a VLA model to generate continuous actions.}
\label{fig:pipeline}
\end{figure}

\textbf{Proposed Method.} A fundamental challenge in this formulation is to identify human intent from interactive gestures. This includes identifying the specified object and determining the corresponding handover response. Such gesture-based intent understanding is largely missing in conventional VLA models. To bridge this gap, we propose \m{} that features a dual-path visual-semantic guidance for modeling human intent, as shown in Fig.~\ref{fig:pipeline}. Specifically, \m{} first performs Visual Gesture Prompting (see Sec.~\ref{subsec:visual_prompting}) to augment visual observations with gesture information $\mathcal{V}$, using hand pose estimation and fingertip-directed rays. The augmented images are then fed into Semantic Intent Parsing (see Sec.~\ref{subsec:semantic_parsing}) to obtain gesture and task descriptions, denoted as $\mathcal{S}$. Finally, the augmented images and the intent reasoning results jointly constitute human intent and are integrated into a VLA model (see Sec.~\ref{subsec:action_generation}) to produce final actions.

\subsection{Visual Gesture Prompting}
\label{subsec:visual_prompting}
VLA models typically rely on detailed textual instructions to specify the target object. However, such explicit verbal specification is inefficient in interactive HRI scenarios, especially when multiple candidate objects appear in the workspace. In contrast, humans more naturally indicate the target through simple pointing gestures, but it is non-trivial for the robot to ground the pointing gesture to the intended object accurately. Particularly when hands are away from the workspace and several objects are spatially close, different targets may correspond to highly similar pointing gestures, and the correct object intention may depend only on subtle differences in the pointing direction. To address this issue, we introduce Visual Gesture Prompting: 

\textbf{Gesture Pointing Ray.} Given the third-person global observation $o_t^{global}$, we employ a hand pose estimator~\cite{pavlakos2024reconstructing} to extract joint keypoints, which provides an explicit representation of gesture intent (e.g., pointing or receiving). However, directly using the estimated hand skeleton is still insufficient for resolving pointing ambiguity. Therefore, when a pointing gesture is detected, we further construct an index-finger pointing ray. Let $p_{base}$, $p_{mid}$, and $p_{top}$ denote the base, middle, and fingertip keypoints of the index finger. We compute the extended middle point and fingertip as:
\begin{equation}
p'_{mid} = K(p_{mid} - p_{base}) + p_{base},
\end{equation}
\begin{equation}
p'_{top} = K(p_{top} - p_{mid}) + p'_{mid},
\end{equation}
where $K$ is the extension factor. As illustrated in Fig.~\ref{fig:pipeline}, the resulting pointing ray extends the local finger direction into the workspace, providing a clearer geometric cue for associating the pointing gesture with the object to be grasped.

\textbf{Visual Overlay.} Unlike previous methods that directly concatenate hand pose embeddings with visual and language tokens, we project the hand skeleton and the pointing ray back onto the 2D image plane and render them as overlaid visual prompts. The augmented observation serves as an explicit visual prompt $\mathcal{V}$, which preserves the original visual structure while encoding the spatial relation between the pointing gesture and candidate objects. This design avoids modifying the pretrained token space of the robot foundation model and does not require additional cross-modal alignment networks. In contrast, naive token concatenation may disrupt pretrained visual-language representations, increase the risk of overfitting to pose locations, and degrade object grasping performance, as validated by the ablation results in Fig.~\ref{fig:fusion_mechanism}.

\subsection{Semantic Intent Parsing}
\label{subsec:semantic_parsing}
While visual gesture prompting improves spatial grounding of manipulation targets, its effect on intent understanding remains limited, as VLA models largely rely on textual instructions to reason about task semantics.
Therefore, the robot must understand the semantic meaning of human gestures to determine the appropriate interaction state (e.g., waiting or executing the handover). Without such intent reasoning, the robot may deliver the object before the human is ready, leading to unsafe or failed interaction.

To address this challenge, we employ a pre-trained VLM~\cite{geminiteam2023gemini} to translate visual gesture cues into high-level semantic guidance and execution instructions. To avoid interrupting the continuous control loop with frequent VLM inference, we introduce a stable-state triggering mechanism. Specifically, the system monitors the estimated pose of the human hand, and the VLM is invoked only when the predicted gesture remains stable for 1.0 second. This query is performed once at the beginning of each task phase. The VLM then parses this stabilized interaction state and outputs a structured semantic tuple $\mathcal{S} = \langle \mathcal{T}_{ges}, \mathcal{T}_{task} \rangle$, where $\mathcal{T}_{ges}$ describes the human gesture and $\mathcal{T}_{task}$ provides robot execution instructions, as described in Fig.~\ref{fig:pipeline}.The generated semantic result is then cached and reused throughout the current phase.

Notably, these textual descriptions are intentionally designed to be \textit{object-agnostic} (i.e., using the generic term ``target object'' instead of specific object names). This design mitigates two potential issues: (1) VLMs may occasionally hallucinate or misclassify object categories, introducing incorrect semantic supervision, and (2) explicit object names cannot resolve spatial ambiguity when multiple visually identical objects exist in the scene. By removing object identity from the semantic pathway, the model relies entirely on explicit visual prompts for precise object grounding. 

\subsection{Action Generation}
\label{subsec:action_generation}

Our continuous control policy, $\pi_\theta$, integrates intent priors with the pre-trained VLA backbone. To avoid introducing learnable modules or modifying the original architecture, we employ a direct latent concatenation strategy. The semantic tuple $\mathcal{S}$ is appended to the text instruction, and the geometric prompt $\mathcal{V}$ is overlaid onto the global observation. The inputs are then processed by the VLM backbone to produce a composite multimodal representation $Z$.

Conditioned on $Z$, the policy generates actions using a Flow Matching objective to regress the vector field of the action distribution:
\begin{equation}
\mathcal{L}_{\mathrm{flow}} = \mathbb{E}_{\tau, \epsilon, a} \left[ \left\| v_\theta(a_\tau, \tau, Z) - (a - \epsilon) \right\|_2^2 \right]
\label{eq:flow_matching}
\end{equation}
where $a$ represents the ground-truth 6-DoF action and gripper state, $\epsilon \sim \mathcal{N}(0, I)$ is Gaussian noise, $\tau \sim \mathcal{U}[0,1]$ is the temporal step, and $a_\tau = (1-\tau)\epsilon + \tau a$ is the interpolated state. Guided by $Z$, the velocity field $v_\theta$ decodes the cross-modal features into final action trajectories.
\section{Experiment}
To systematically analyze the proposed \m{} approach, we design experiments to answer the following core questions: 
\textbf{Q1.} How do the visual and semantic pathways contribute to overall performance? 
\textbf{Q2.} How does visual gesture prompting resolve spatial ambiguity that purely semantic reasoning struggles with, thereby justifying our dual-path guidance design?
\textbf{Q3.} How robust is the approach under unseen spatial layouts? 
\textbf{Q4.} How well does the method generalize to unseen human participants with varying physical characteristics?

\subsection{Experimental Details}

\begin{wrapfigure}{r}{0.55\textwidth}
\vspace{-10pt}
\centering
\includegraphics[width=\linewidth]{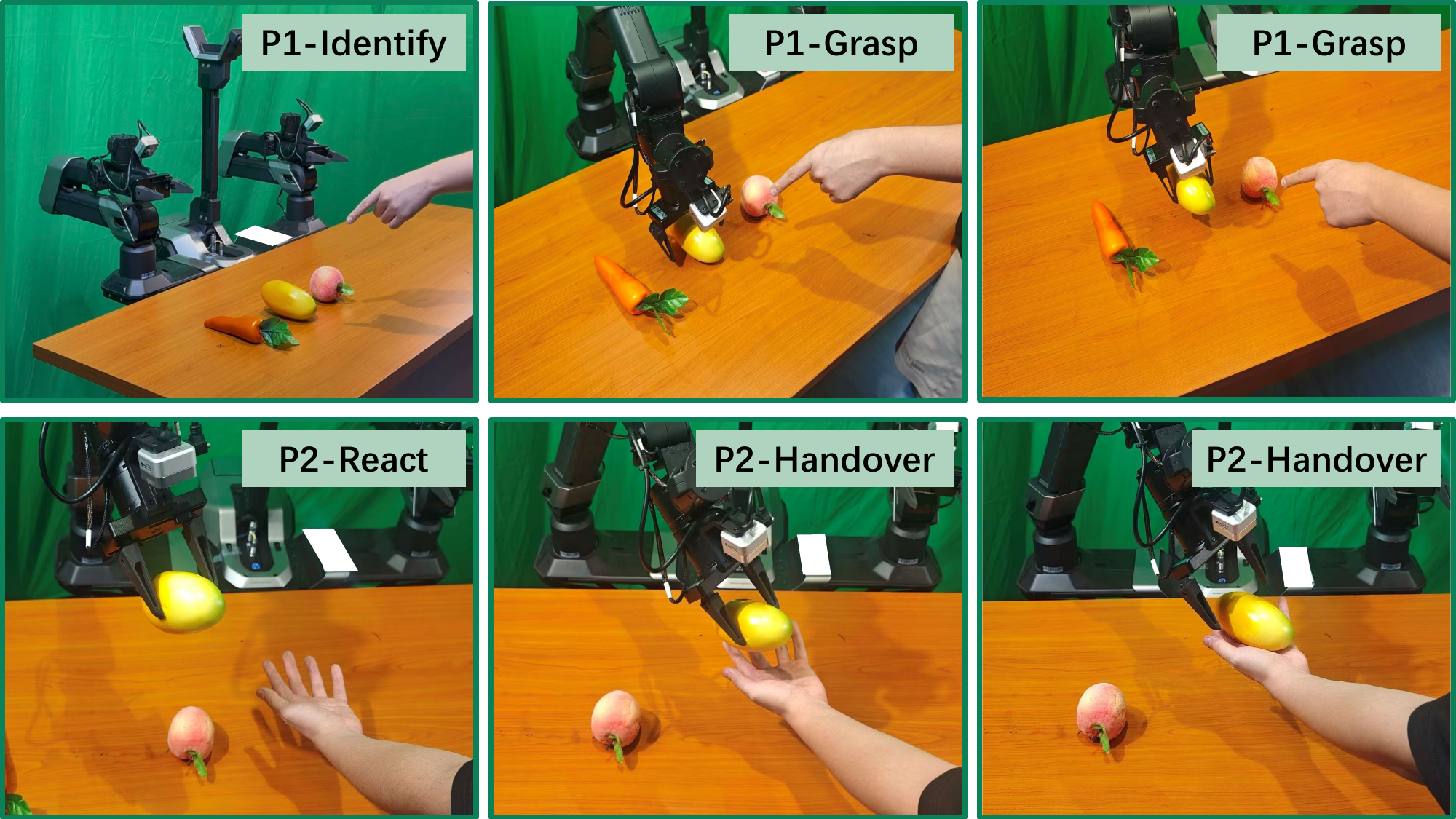}
\caption{Illustration of the two-phase HRI task.}
\label{fig:handover_main}
\end{wrapfigure}

\textbf{Setup.} As illustrated in Fig.~\ref{fig:handover_main}, we evaluate our approach on a sequential human-robot interaction (HRI) workflow comprising two distinct phases: (i) \textbf{Phase 1 (Pointing \& Picking):} A human specifies a target object among multiple items via a pointing gesture. The robot is then required to decode this spatial intent, identify the designated item, and execute a stable grasp. (ii) \textbf{Phase 2 (Response \& Handover):} Upon a successful grasp, the robot remains stationary while monitoring human states. Once the human relocates their hand and transitions to an ``open palm'' gesture, the robot needs to plan a trajectory above the palm, performing a precise handover and releasing the object. All physical experiments are conducted on a Galaxea R1-Lite wheeled robot with two 6-DoF arms. 
Dual-perspective observations are utilized: a body-mounted camera captures global third-person views, while a wrist-mounted camera provides egocentric perspectives.
To evaluate spatial robustness and mitigate overfitting, the environment incorporates three target objects (i.e., peach, mango, and carrot) tested across three distinct locations.

\textbf{Implementation Details.} The policy network is built upon the pre-trained $\pi_{0.5}$ model.
Since dynamic human-robot interaction is underrepresented in the original pre-training distribution, we perform full-parameter fine-tuning on the multi-modal input sequences. This comprehensive adaptation is crucial to seamlessly adapt the pre-trained representations to the HRI domain and ensure deep cross-modal alignment between the injected visual/textual priors and continuous action generation. 
The model is trained on a dataset of 270 real-world human-robot demonstration episodes using the AdamW optimizer with a learning rate of $1 \times 10^{-8}$ and a global batch size of 64. The training process takes approximately 26 hour with 4 $\times$ NVIDIA A800 GPUs.

We evaluate the performance using four conditional success rates (SR) to provide a more fine-grained evaluation. For Phase 1, we measure \textbf{Identify SR} (the robot correctly moving toward the designated target) and \textbf{Grasp SR} (achieving a stable grasp given correct identification).
For Phase 2, we measure \textbf{React SR} (the robot initiating movement after the gesture switch) and \textbf{Handover SR} (accurate delivery and release above the palm given a successful response).

\subsection{Results and Analysis}
\textbf{Performance \& Ablation.} Addressing \textbf{Q1}, we first evaluate the overall execution performance of our end-to-end system under the sequential grasp-then-handover tasks. As detailed in Table \ref{tab:fine_grained_simple}, the pure end-to-end baseline exhibits significant performance degradation during continuous execution. Without explicit visual guidance in Phase 1, the baseline achieves only 46.7\% Identify SR, which further collapses to 6.7\% Grasp SR, reflecting unreliable target localization. 
In contrast, by deeply integrating visual geometric cues with detailed semantic instructions, \m{} achieves a strong 86.7\% Identify SR and consistently maintains 80.0\% success across all subsequent execution stages (Grasp, React, and Handover), demonstrating robust performance in dynamic HRI settings.

\begin{wrapfigure}{r}{0.55\textwidth}
\vspace{-10pt}
\centering
\includegraphics[width=\linewidth]{"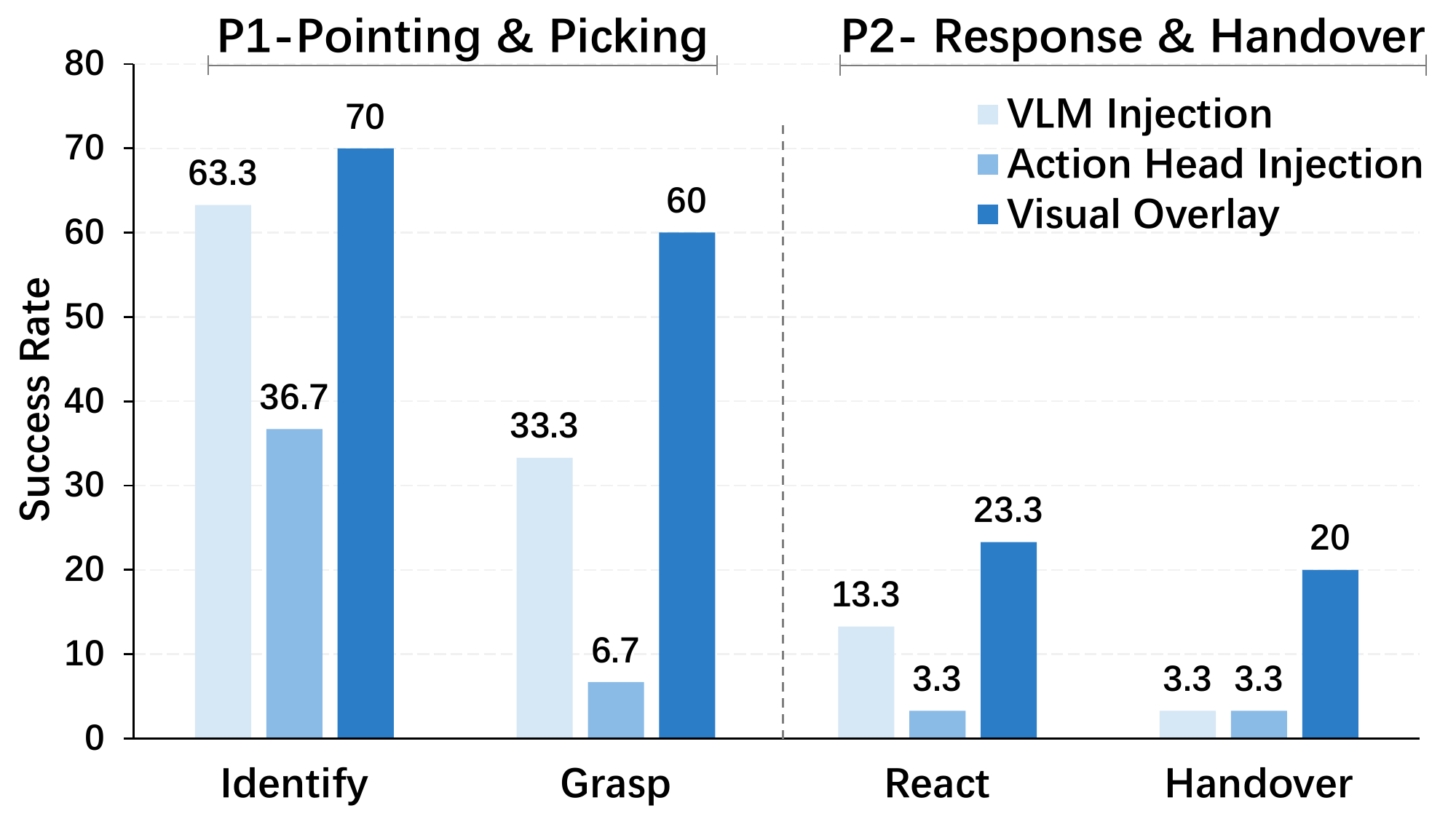"}
\caption{Ablation results on gesture keypoint integration strategies.}
\label{fig:fusion_mechanism}
\end{wrapfigure}

To understand the source of these gains, we conduct a detailed ablation study. We first analyze different strategies for utilizing hand gesture information in Fig.~\ref{fig:fusion_mechanism}, focusing on how gesture cues are incorporated into the policy learning. 
We find that our simplest parameter-free visual overlay (skeleton and fingertip rays) achieves the best overall performance, significantly improving early-stage grounding with 70.0\% Identify SR.
In comparison, token-based variants introduce additional encoding modules to transform 2D keypoints into latent representations. VLM-base token injection improves identification ability (63.3\%) but loses geometric precision in downstream manipulation (Grasp SR: 33.3\%, Handover SR: 3.3\%), while injecting tokens into the action head performs even worse due to degraded grounding capacity (Identify SR: 36.7\%, Grasp SR: 6.7\%). 
These results suggest that direct visual overlay can preserve pretrained representations, whereas extra tokenization may disrupt feature alignment.

We further study the role of each design by comparing visual-only and full visual-semantic guidance. 
While visual guidance alone improves performance over the baseline, purely keypoint-based geometric prompting provides only limited benefits.
In contrast, explicitly introducing fingertip-directed rays to indicate the target object leads to a substantial improvement in early-stage grounding.
However, visual guidance remains insufficient for handling temporal state transitions, with notable degradation in reactive and handover stages (23.3\% React SR). In contrast, the full model combining both visual and semantic priors provides complementary improvements, enabling consistent high performance (80\% Grasp, React, and Handover SR). 

\begin{table}[t]
\centering
\renewcommand{\arraystretch}{1.15} 
\caption{\textbf{Fine-grained success rate comparison.} Detailed execution performance of the Baseline and the proposed \m{} across different spatial settings.}
\label{tab:fine_grained_simple}
\vspace{1mm}
\small 
\begin{tabular}{@{} ll cccc @{}} 
\toprule
\textbf{Method} & \textbf{Setting} & \textbf{Identify SR} & \textbf{Grasp SR} & \textbf{React SR} & \textbf{Handover SR} \\
\midrule

\multirow{3}{*}{$\pi_{0.5}$}
 & Setting 1 & 6/15 (40.0\%) & 2/15 (13.3\%) & 1/15 (6.7\%) & 0/15 (0.0\%) \\
 & Setting 2 & 8/15 (53.3\%) & 0/15 (0.0\%) & 0/15 (0.0\%) & 0/15 (0.0\%) \\
 \cmidrule(lr){2-6}
 & Overall (Avg) & 14/30 (46.7\%) & 2/30 (6.7\%) & 1/30 (3.3\%) & 0/30 (0.0\%) \\
\midrule

\multirow{3}{*}{\m{}} 
 & Setting 1 & 12/15 (80.0\%) & 11/15 (73.3\%) & 11/15 (73.3\%) & 11/15 (73.3\%) \\
 & Setting 2 & 14/15 (93.3\%) & 13/15 (86.7\%) & 13/15 (86.7\%) & 13/15 (86.7\%) \\
 \cmidrule(lr){2-6}
 & Overall (Avg) & \textbf{26/30 (86.7\%)} & \textbf{24/30 (80.0\%)} & \textbf{24/30 (80.0\%)} & \textbf{24/30 (80.0\%)} \\
 
\bottomrule
\end{tabular}
\vspace{-6pt}
\end{table}

\begin{table}[t]
\centering
\renewcommand{\arraystretch}{1.15} 
\caption{\textbf{Ablation on visual and semantic guidance.} Overall average success rates across all spatial settings. VGP denotes Visual Geometric Prompting, and SIP indicates Semantic Intent Parsing.}
\label{tab:fine_grained_ablation}
\vspace{1mm}
\small
\begin{tabular}{@{} l cccc @{}}
\toprule
\textbf{Method} & \textbf{Identify SR} & \textbf{Grasp SR} & \textbf{React SR} & \textbf{Handover SR} \\
\midrule
$\pi_{0.5}$ & 46.7\% & 6.7\% & 3.3\% & 0.0\% \\
+ VGP (keypoints) & 56.7\% & 30.0\% & 13.3\% & 3.3\% \\
+ VGP (keypoints + ray) & 70.0\% & 60.0\% & 23.3\% & 20.0\% \\
+ VGP \& SIP & \phantom{00}\textbf{86.7\%} \textsubscript{\textcolor{blue}{$\uparrow$40.0\%}} & \phantom{00}\textbf{80.0\%}\textsubscript{\textcolor{blue}{$\uparrow$73.3\%}} & \phantom{00}\textbf{80.0\%}\textsubscript{\textcolor{blue}{$\uparrow$76.7\%}} & \phantom{00}\textbf{80.0\%}\textsubscript{\textcolor{blue}{$\uparrow$80.0\%}} \\
\bottomrule
\end{tabular}
\end{table}

\textbf{Rationale for Visual-Semantic Guidance.} To answer \textbf{Q2}, we employ a pre-trained VLM as a probe to assess the necessity of explicit visual prompting for intent grounding. Using a structured prompt template (Appendix A.2), we define three metrics to evaluate the semantic parsing outputs across 20 independent trials: (1) \textbf{Gesture Recognition} (identifying intents like pointing or open-palm gestures); (2) \textbf{Target Grounding} (accurately associating the gesture with the intended object among candidates); and (3) \textbf{Execution Suggestion} (generating correct task instructions). 

As shown in Table \ref{tab:perception_validation}, using raw RGB images is sufficient for recognizing gesture states and basic execution logic, achieving accuracy comparable to that with visual overlay. However, it performs poorly on spatial grounding, yielding only a 40.0\% Target Grounding accuracy. In contrast, rendering explicit geometric cues onto the 2D observation space effectively resolves this spatial ambiguity, boosting the target grounding accuracy to 90.0\%. 
These findings justify our visual-semantic guidance design: VLM excels at high-level task phase reasoning, but is weak in precise spatial anchoring. Therefore, in our actual pipeline, the Desired Interactive Target field is intentionally abstracted away to maintain the object-agnostic design detailed in Section \ref{subsec:semantic_parsing}. By explicitly decoupling these roles, we leverage the semantic pathway exclusively for robust task state description, while delegating spatial target grounding entirely to the visual pathway.

\begin{table}[t]
\centering
\renewcommand{\arraystretch}{1.15}
\caption{\textbf{Validation of visual-guided semantic reasoning.} Accuracy across three metrics over 20 independent trials.}
\label{tab:perception_validation}
\vspace{1mm}
\resizebox{\textwidth}{!}{
\begin{tabular}{@{} l l cc @{}}
\toprule
\textbf{Evaluation Metric} & \textbf{Evaluated Output Fields} & \textbf{w/o Visual Overlay} & \textbf{w/ Visual Overlay} \\
\midrule
\textbf{Gesture Recognition} & \texttt{[Human Gesture Description]} & 75.0\% & \textbf{80.0\%} \\
\textbf{Target Grounding} & \texttt{[Desired Interactive Target]} & 40.0\% & \textbf{90.0\%} \\
\textbf{Execution Suggestion} & \texttt{[Robotic Arm Execution Instruction]} & 80.0\% & 80.0\% \\
\bottomrule
\end{tabular}
}
\vspace{-6pt}
\end{table}

\textbf{Robustness \& Generalization.} To answer \textbf{Q3} and \textbf{Q4}, we comprehensively evaluate the robustness and generalization capability of our approach under unseen spatial layouts and across unseen human participants with diverse body shapes and clothing. 
Table~\ref{tab:overall_robustness} reports the success rates of the initial stage (Identify \& Grasp) across unseen spatial configurations and diverse interactive objects. The baseline model exhibits significant performance degradation and high variance across object locations, indicating sensitivity to spatial distribution shifts. Conversely, our proposed integrated strategy maintains higher and more stable success rate across all challenging setups (Locations A, B, and C with different objects). 

Furthermore, we evaluate the Identify SR across three subjects (Table~\ref{tab:unseen_human}): one participant seen during training, and two completely unseen participants. These unseen individuals feature hand sizes, body shapes, and clothing styles distinct from the training distribution. The baseline's identification capability deteriorates significantly when interacting with unseen humans, suggesting limited robustness to variations in physical scale and appearance. In contrast, \m{} achieves consistent identification success rate across all subjects. 
This highlights an architectural advantage of our design: the hand pose estimator provides an explicit hand skeleton for visual guidance, decoupling intent reasoning from visual appearance and environmental variations.

\begin{table}[t]
\centering
\small 

\begin{minipage}[t]{0.56\textwidth}
    \centering
    \setlength{\tabcolsep}{3pt} 
    \renewcommand{\arraystretch}{1.15}
    \caption{\textbf{Robustness across different spatial layouts and interactive objects.} Success rates are reported for the initial Identify \& Grasp stage.}
    \label{tab:overall_robustness}
    \vspace{1mm}
    \begin{tabular}{@{} l cc cc cc @{}}
    \toprule
    \multirow{2}{*}{\textbf{Location}} & \multicolumn{2}{c}{\textbf{Peach}} & \multicolumn{2}{c}{\textbf{Mango}} & \multicolumn{2}{c}{\textbf{Carrot}} \\
    \cmidrule(lr){2-3} \cmidrule(lr){4-5} \cmidrule(lr){6-7}
     & $\pi_{0.5}$ & \m{} & $\pi_{0.5}$ & \m{} & $\pi_{0.5}$ & \m{} \\
    \midrule
    A & $5/5$ & $4/5$ & $3/5$ & $5/5$ & $1/5$ & $4/5$ \\
    B & $3/5$ & $5/5$ & $3/5$ & $5/5$ & $0/5$ & $5/5$ \\
    C & $1/5$ & $3/5$ & $2/5$ & $4/5$ & $1/5$ & $5/5$ \\
    \bottomrule
    \end{tabular}
\end{minipage}
\hfill
\begin{minipage}[t]{0.42\textwidth}
    \centering
    \setlength{\tabcolsep}{7pt} 
    \renewcommand{\arraystretch}{1.15}
    \caption{\textbf{Generalization to unseen human participants.} Evaluated strictly on the Identify SR.} 
    \label{tab:unseen_human}
    \vspace{1mm}
    \begin{tabular}{@{} l c cc @{}}
    \toprule
    \multirow{2}{*}{\textbf{Method}} & \multirow{2}{*}{\textbf{Seen Pt.}} & \multicolumn{2}{c}{\textbf{Unseen}} \\
    \cmidrule(l){3-4}
     & & \textbf{Pt. A} & \textbf{Pt. B} \\
    \midrule
    $\pi_{0.5}$ & 4/10 & 1/10 & 0/10 \\
    \m{} & \textbf{8/10} & \textbf{8/10} & \textbf{6/10} \\
    \bottomrule
    \end{tabular}
\end{minipage}
\end{table}

\section{Limitations}
\label{sec:limitations}
Despite its effectiveness, our method has several limitations that suggest directions for future research. First, visual gesture prompting depends on the accuracy of the hand pose estimator, and may degrade in challenging cases such as severe motion blur or occlusions. Second, the current pointing ray is generated using a fixed extension scalar $K$, which is not adaptive to scene depth and may be suboptimal in environments with large depth variations.
Future work will focus on improving robust gesture perception under occlusion and developing adaptive ray mechanisms.

\section{Conclusion}
\label{sec:conclusion}

In this work, we propose \m{} that enhances VLA models with human gesture understanding for more reliable human-robot interaction. 
\m{} introduces a dual-path design that injects complementary guidance into frozen VLA policies: a visual pathway that provides geometric grounding through gesture-aware visual overlays, and a semantic pathway that offers high-level textual descriptions of human intent and task context.
Without modifying the policy architecture, this design enables accurate alignment between human gestures and robot actions. Real-world experiments demonstrate that \m{} consistently improves robustness and generalization across unseen spatial layouts and diverse human partners. Overall, our results suggest that incorporating gesture-based multimodal intent cues is a simple yet effective way to enhance VLA-based human-robot interaction.

\clearpage
\newpage
\bibliographystyle{assets/plainnat}
\bibliography{paper}

\clearpage
\newpage
\appendix
\section*{Appendix}
\appendix

\section{Additional Details}

\subsection{Timing and Latency Analysis}
We present a detailed timing analysis of the proposed \m{} framework during real-world deployment, highlighting the computational overhead introduced by the visual and semantic guidance pathways. As shown in Table \ref{tab:timing}, the visual gesture prompting step (which involves capturing the third-person RGB image, executing the hand pose estimator to extract 2D keypoints, and computing the pointing ray) requires approximately $7.35$ ms. This pre-processing is crucial for generating the explicit visual prompt ($\mathcal{V}$).


To prevent inference latency from bottlenecking the high-frequency continuous control, the semantic extraction via the VLM is executed asynchronously and strictly at the initial inference step of each task phase. During this phase-initial step, querying the Gemini API to generate the explicit semantic caption takes roughly $4.65$ seconds. Once the multi-modal prompts are constructed, the forward pass of the $\pi_{0.5}$ base model to generate continuous action chunks operates at approximately $3.42$ s. The major source of phase-initial latency stems from the API-based VLM query, which could be mitigated in future iterations by deploying a distilled, local VLM. However, because this query is strictly triggered only once at the beginning of each new task phase, it does not block the high-frequency control loop, ensuring that the actual physical manipulation remains smooth and uninterrupted.

\begin{table}[h]
\centering
\renewcommand{\arraystretch}{1.15}
\caption{\textbf{Timing of each component in the proposed method.} }
\label{tab:timing}
\resizebox{\textwidth}{!}{
\begin{tabular}{@{} l c c c c @{}}
\toprule
 & \textbf{Visual Prompting} & \textbf{Semantic Parsing} & \textbf{Policy Inference} & \textbf{Control Execution} \\
 & \textit{(Hand Pose Estimator)} & \textit{(Once per phase)} & \textit{($\pi_{0.5}$)} & \\
\midrule
\textbf{Time} & $7.35 \pm 2.67$ ms & $4.65 \pm 1.36$ s & $3.42 \pm 0.35$ s & $3.23 \pm 0.06$ s \\
\bottomrule
\end{tabular}
}
\end{table}

\subsection{VLM Prompt Design for Intention Perception}
To standardize the extraction of semantic features during these phase-initial queries, we prompt the Gemini 2.0 Flash API for semantic intent parsing using a strict template. The translation of the prompt used in our experiments is provided below:

\vspace{1em} 

\begin{tcolorbox}[
    colback=gray!5!white,        
    colframe=gray!40!black,      
    arc=2mm,                     
    boxrule=0.5pt,               
    left=10pt, right=10pt, top=8pt, bottom=8pt, 
    title=\textbf{System Prompt Template for Visual Intention Perception}, 
    coltitle=black,              
    colbacktitle=gray!15!white,  
    fontupper=\small\rmfamily    
]

\textbf{System Prompt:} You are a professional embodied AI semantic reasoning assistant. Please analyze the human actions in these two images and infer their intentions. Please reply strictly according to the following template, without outputting any extraneous text: \\

\textbf{[Human Gesture Description]:} \textit{\textless Detailed description of the key gesture in the image, e.g., The human hand is pointing at target object. / The human hand is keeping open for receiving the target object. etc.\textgreater} \\[0.5em]

\textbf{[Interactive Target]:} \textit{\textless The specific object being manipulated, e.g., The target object is a (peach, mango, carrot)\textgreater} \\[0.5em]

\textbf{[Robotic Arm Execution Suggestion]:} \textit{\textless Based on the human's current action, what strategy should the collaborative robotic arm adopt? e.g., Pick up the target object and wait for human hand changes to next gesture. / Handover the target object to the human hand when human hand changes gesture. etc.\textgreater}
\end{tcolorbox}

\subsection{Dataset Collection and Processing}
\label{subsec:appendix_dataset}

To train the continuous control policy, we collected real-world demonstration data via a teleoperation system on the Galaxea R1-Lite robot. Aligning with our task formulation (Section 4.1), each complete human-robot interaction episode was structurally recorded as two sequential data segments: Phase 1 (Pointing \& Picking) and Phase 2 (Response \& Handover). 

Crucially, to support our dual-pathway guidance design, the recorded data for each phase was individually processed and structured. For the visual pathway, the geometric prompts (hand skeletons and fingertip-directed rays) were explicitly rendered and overlaid onto the global observations for each frame within the segment. For the semantic pathway, textual instructions were assigned separately for each phase to reflect the distinct interaction states. Following the object-agnostic principle established in Section 3.3, these textual prompts provide high-level execution directives (e.g., ``pick up the target object and wait'' for Phase 1, and ``handover the object to the human'' for Phase 2) while intentionally abstracting away specific object names. 

This independent, phase-specific collection and processing strategy ensures that the VLA policy learns to transition between dynamic interaction states based on semantic cues, while relying entirely on the explicitly constructed visual prompts for spatial target grounding.

\section{Visual Comparison: Keypoints vs. Ray Integration}
\label{subsec:appendix_ray_ablation}

\begin{figure}[htbp]
\centering
\begin{subfigure}{0.48\textwidth}
    \centering
    \includegraphics[width=\linewidth]{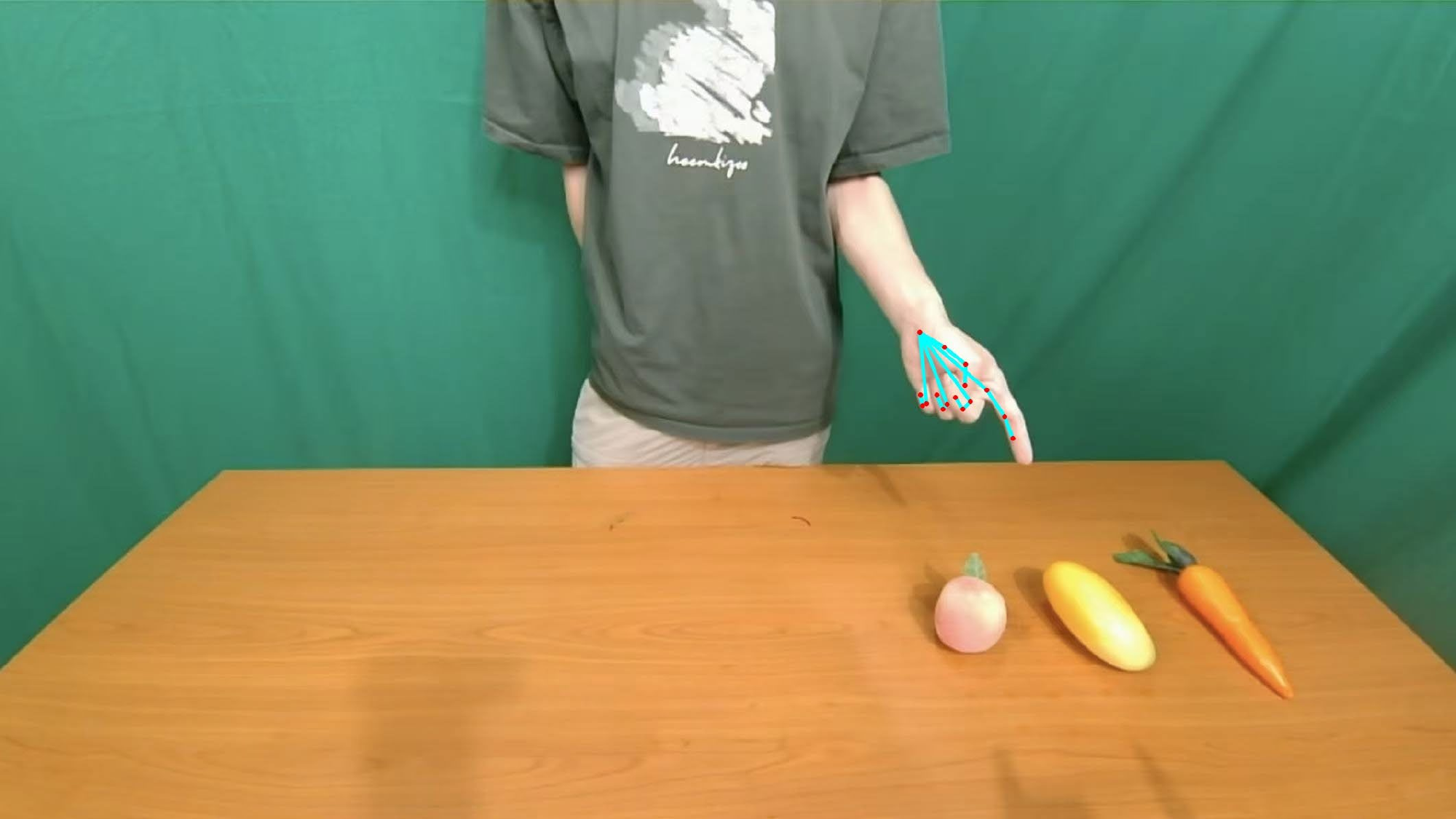} 
    \caption{Visual Prompt with Keypoints Only}
    \label{fig:keypoints_only}
\end{subfigure}
\hfill
\begin{subfigure}{0.48\textwidth}
    \centering
    \includegraphics[width=\linewidth]{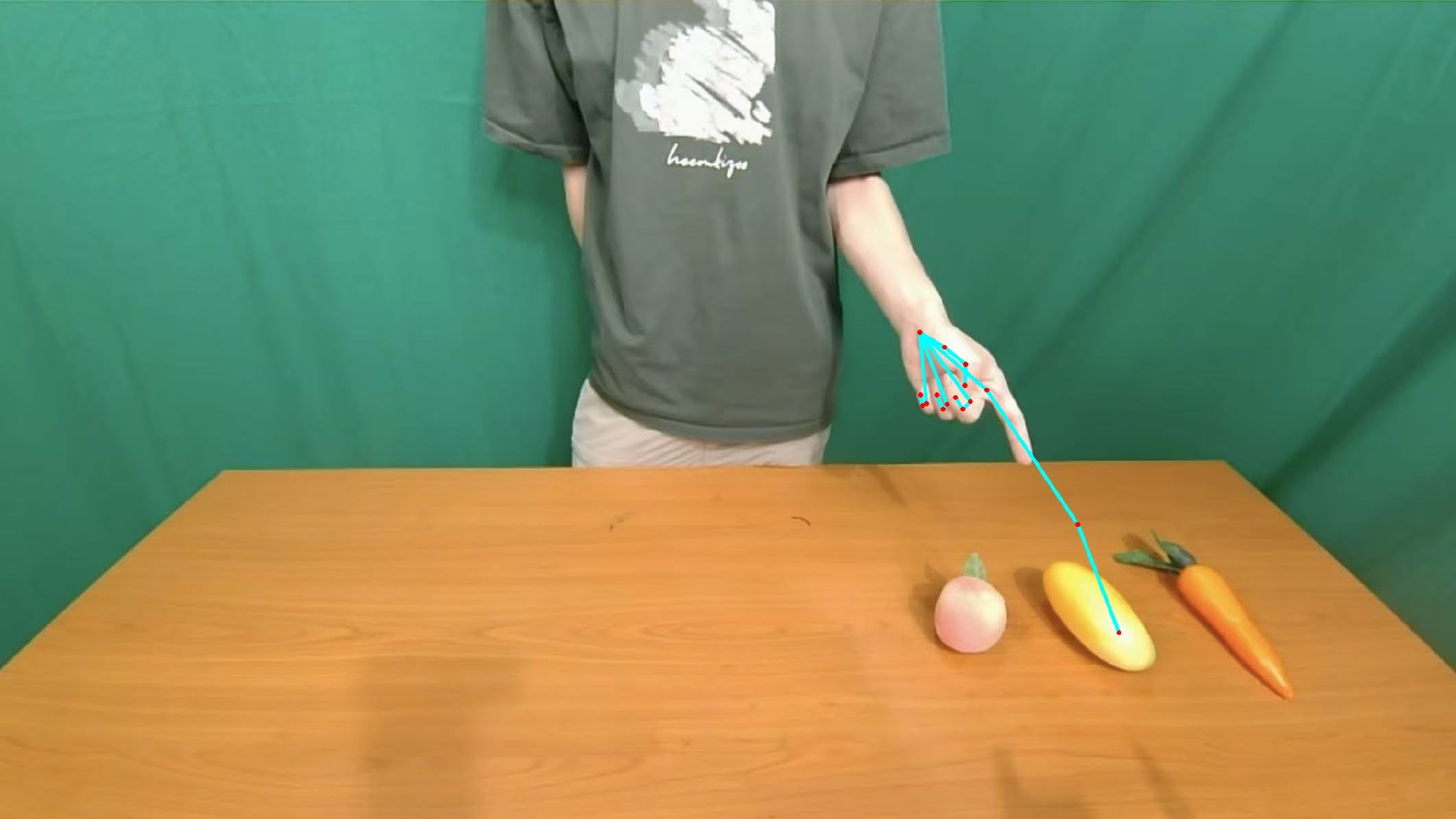} 
    \caption{Visual Prompt with Keypoints + Ray}
    \label{fig:keypoints_ray}
\end{subfigure}
\caption{\textbf{Qualitative comparison of visual geometric prompts.} (a) Relying purely on keypoints leaves the precise target object ambiguous when candidate items are grouped closely. (b) Augmenting the keypoints with an explicit pointing ray directly anchors the intended target.}
\label{fig:appendix_ray_comparison}
\end{figure}

To provide deeper intuition for the ablation results presented in Table~\ref{tab:fine_grained_ablation} (Section 4.2), we visually compare the two geometric prompting strategies in Fig.~\ref{fig:appendix_ray_comparison}. 

In scenarios where the human hand operates at a distance from the candidate objects, relying solely on hand keypoints (Fig.~\ref{fig:appendix_ray_comparison}a) captures the local gesture but leaves the exact spatial target ambiguous. The VLA model is forced to implicitly extrapolate the pointing direction across the 3D workspace, which often leads to misidentification when multiple visually similar or clustered objects (e.g., the peach, mango, and carrot) are present. This inherent spatial ambiguity explains the limited early-stage grounding performance (56.7\% Identify SR) observed in the purely keypoint-based configuration.

In contrast, as shown in Fig.~\ref{fig:appendix_ray_comparison}b, explicitly computing and rendering the fingertip-directed ray directly bridges the spatial gap between the human gesture and the physical target. By extending the index finger's orientation vector onto the 2D observation plane, the visual prompt explicitly intersects with or points directly at the intended target (in this case, the mango). This explicit geometric anchoring drastically reduces the spatial reasoning burden on the pre-trained foundation model, effectively resolving target ambiguity and driving the significant performance leap in early-stage identification (70.0\% Identify SR).

\section{Failure Case Analysis: Limitations of Fixed Ray Extension}
\label{subsec:appendix_failure_cases}

\begin{figure}[htbp]
\centering
\begin{subfigure}{0.32\textwidth}
    \centering
    \includegraphics[width=\linewidth]{"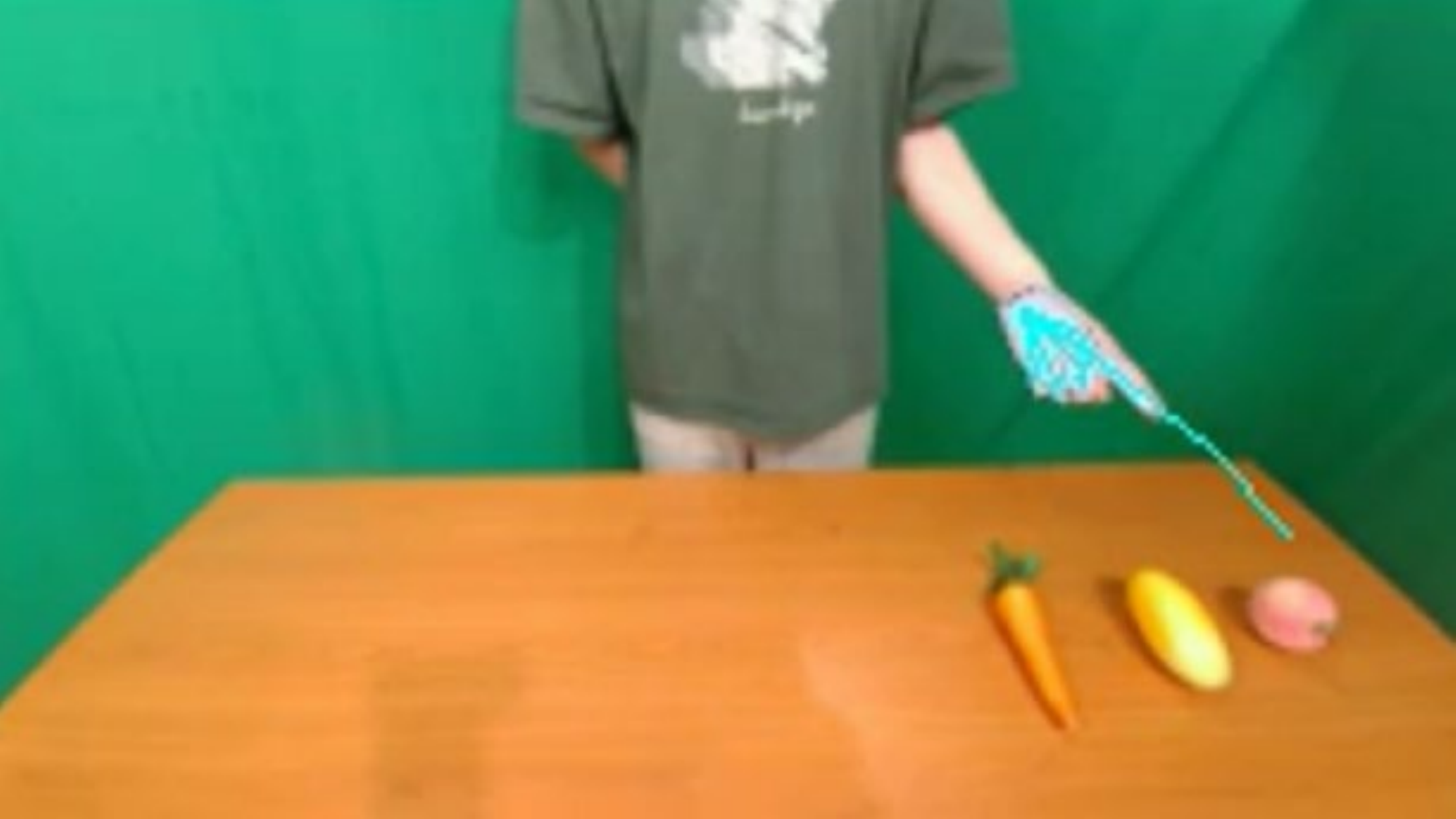"}
    \caption{Ambiguous Pointing Ray}
    \label{fig:failure_ambiguous}
\end{subfigure}
\hfill
\begin{subfigure}{0.32\textwidth}
    \centering
    \includegraphics[width=\linewidth]{"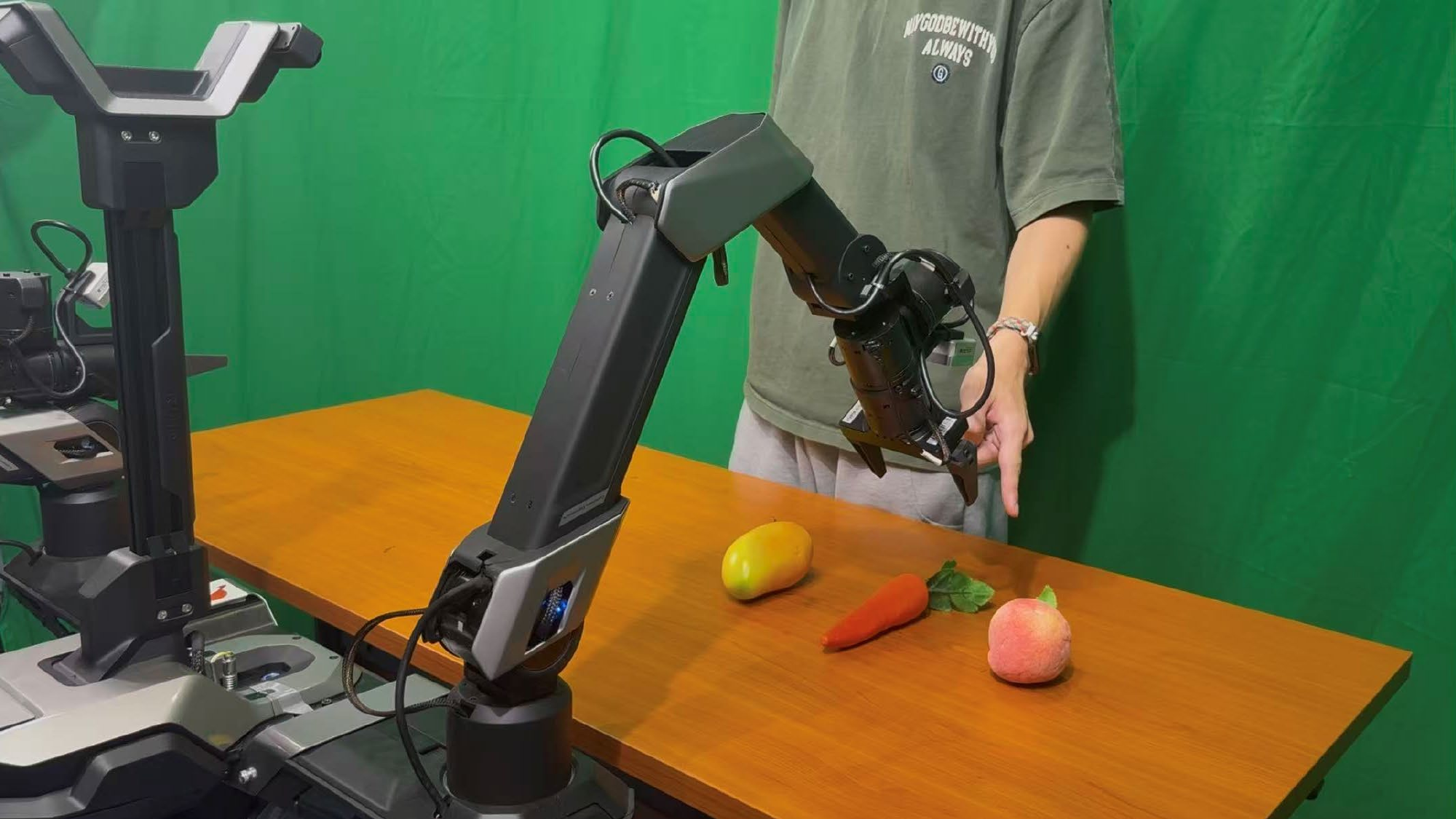"}
    \caption{Wrong Target Selection}
    \label{fig:failure_wrong}
\end{subfigure}
\hfill
\begin{subfigure}{0.32\textwidth}
    \centering
    \includegraphics[width=\linewidth]{"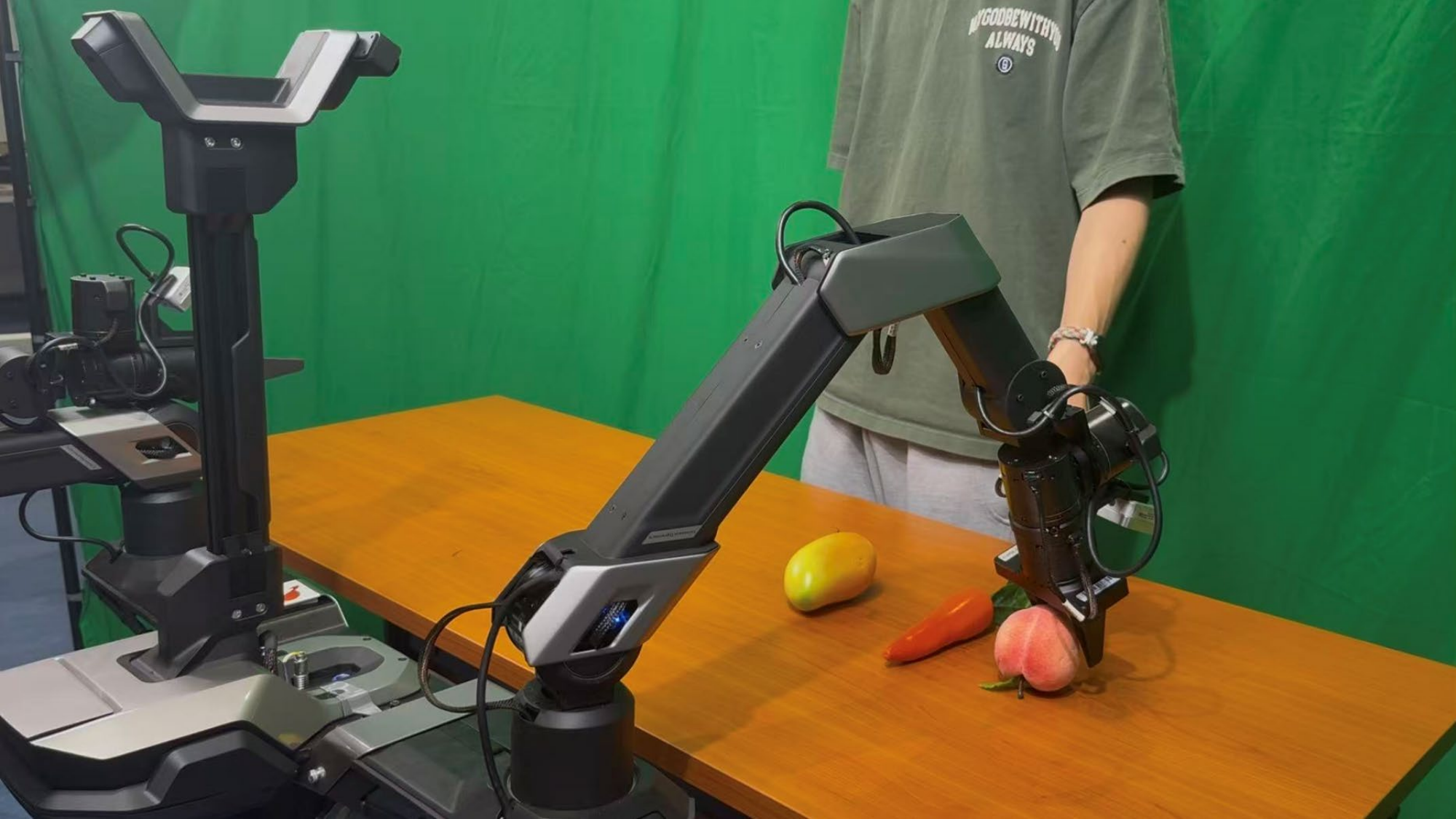"}
    \caption{Grasp Missing}
    \label{fig:failure_miss}
\end{subfigure}
\caption{\textbf{Failure cases in extreme distant interactions.} When the human hand operates exceptionally far from the workspace, the 2D ray can become ambiguous due to the lack of depth information (a), which occasionally leads to wrong target selection (b) or grasp failures (c).}
\label{fig:failure_cases}
\end{figure}

While the proposed \m{} framework handles typical interaction distances effectively, failures can occasionally occur in extreme edge cases where the human hand is positioned exceptionally far from the workspace. Due to the fixed ray extension scalar $K$ and the lack of explicit depth information, the 2D projected ray might point ambiguously between candidate items (Fig.~\ref{fig:failure_cases}a). This spatial ambiguity can confuse the downstream VLA policy, leading to incorrect target selection (Fig.~\ref{fig:failure_cases}b) or grasp misses (Fig.~\ref{fig:failure_cases}c).

Importantly, these failure modes do not imply that the geometric prompt must perfectly intersect the target to succeed. Leveraging its learned spatial reasoning, the policy exhibits robustness to imperfect geometric prompts, grounding human intent even if the 2D pointing ray fails to directly intersect the target. Nevertheless, to address the aforementioned extreme scenarios, exploring depth-adaptive ray mechanisms remains a valuable direction for future work.



\end{document}